\documentclass[10pt,journal]{IEEEtran}

\ifCLASSOPTIONcompsoc
\usepackage[nocompress]{cite}
\else
\usepackage{cite}
\fi

\ifCLASSINFOpdf

\else

\fi

\usepackage{amssymb}
\usepackage{graphicx}
\usepackage{subfig}
\usepackage{float}
\usepackage{amsmath}
\usepackage{booktabs}
\usepackage{url}
\usepackage[numbers,sort&compress]{natbib}
\usepackage{array}
\usepackage{multicol, blindtext}
\usepackage{lipsum}
\usepackage{ragged2e}
\usepackage[linesnumbered, ruled, vlined]{algorithm2e}

\usepackage{algpseudocode}

\begin{document}
\title{Modeling Historical AIS Data  For Vessel Path Prediction: A Comprehensive Treatment}
\author{Enmei Tu,
        Guanghao Zhang,
		Shangbo Mao,
		Lily Rachmawati,
        Guang-Bin Huang, ~\IEEEmembership{Senior Member,~IEEE}
\thanks{Enmei Tu is with School of Electronics, Information and Electrical Engineering, Shanghai Jiao Tong University, China}
\thanks{Guanghao Zhang, Shangbo Mao and Guang-Bin Huang are with School of Electrical \& Electronic Engineering, Nanyang Technological University, Singapore}
\thanks{Lily Rachmawati is with Computational Engineering Team, Advanced Technology Centre, Rolls-Royce Singapore Pte Ltd}}
\markboth{}%
{Tu \MakeLowercase{\textit{et al.}}: AIS Data For Intelligent Navigation}

\IEEEtitleabstractindextext{

\begin{abstract}
The prosperity of artificial intelligence has aroused intensive interests in intelligent/autonomous navigation, in which path prediction is a key functionality for decision supports,  e.g. route planning, collision warning, and traffic regulation. For maritime intelligence, Automatic Identification System (AIS) plays an important role because it recently has been made compulsory for large international commercial vessels and is able to provide nearly real-time information of the vessel. Therefore AIS data based vessel path prediction is a promising way in future maritime intelligence.
However,  real-world AIS data collected online are just highly irregular trajectory segments (AIS message sequences) from different types of vessels and geographical regions, with possibly very low data quality. So even there are some works studying how to build a path prediction model using  historical AIS data, but still, it is a very challenging problem. In this paper,  we propose a comprehensive framework to model massive historical AIS trajectory segments for accurate vessel path prediction.  Experimental comparisons with existing popular methods are made to validate the proposed approach and results show that our approach could outperform the baseline methods by a wide margin. 
\end{abstract}

}

\maketitle
\IEEEdisplaynontitleabstractindextext
\IEEEpeerreviewmaketitle

\section{Introduction}

 \IEEEPARstart{W}{ith} the rapid globalization and increasing demand for maritime transportation over the world, maritime safety has attracted huge attention in both marine and trading sectors during the past decade. Statistics show that more than 90\% of the world's trade is transported by sea nowadays\footnote{UN-Bussiness Action Hub: https://business.un.org/en/entities/13}, but meanwhile, a significant part of maritime accidents  (about 80\%) have been caused by human errors \cite{wiegmann2017human, kim2011case}.  These accidents have caused severe people  casualties,  economic losses, and environmental crisis. Therefore, the importance of maritime navigation safety has been raised to an unprecedented level in the marine industry \cite{knapp2010comprehensive}.   
 
 Maritime intelligent/autonomous systems can provide potential solutions in the future to improve safety during navigation and management \cite{statheros2008autonomous, alessandrini2018estimated}. For this purpose, Automatic Identification System (AIS) recently has been made compulsory for international/regional commercial vessels above a certain tonnage, including cargo, passage, tankers, etc, as well as parts of civil vessels, including fishing ships, lifeboats, and leisure vessels \cite{harre2000ais}. Comparing with traditional maritime equipment such as radar, sonar or closed-circuit television (CCTV), AIS has many advantages. AIS messages provide rich information of its host ship in a nearly real-time way. Furthermore, AIS messages can be transmitted to and received from a very long distance (20 nautical miles for onboard transceivers and hundreds of nautical miles for satellite receivers \cite{cervera2011satellite}). Moreover, AIS is less likely to be affected by external factors such as sea conditions and weather conditions. A large amount of AIS data are collected every day from different vessels at different locations and the gathered data contain a wealth of information useful for maritime safety, security, and efficiency promotion. However, as will be elaborated in the following section, AIS data usually exhibits many defects, such as low data quality, highly irregular between-message time intervals and poor data integrity \cite{harati2007automatic, last2014comprehensive}. Plus trajectory diversity (different types of vessel, different geographical contexts and different maneuvering statues), therefore the problem of utilizing  massive historical AIS data to build an efficient model for better vessel path prediction is a  challenging task and needs more effort.

 Recently, there have been several attempts trying to utilize big historical AIS data to improve vessel path prediction, but these works have some restrictions or limitations, e.g. requiring  prior knowledge or good data presence. Here we mention a few representative works. For more details, we refer readers to \cite{tu2017exploiting}. Pallotta \textit{et al.} \cite{pallotta2014context} propose an interesting Ornstein-Uhlenbeck Processes based vessel position prediction algorithm, but it  relies on externally extracted contextual information, i.e. which assumes that a vessel must follow the route provided by the Traffic Route Extraction for Anomaly Detection (TREAD) tool. Mazzarella \textit{et al.} \cite{mazzarella2015knowledge} develop an effective Bayesian vessel prediction approach based on a Particle Filter, but it requires prior knowledge from marine traffic analysis. Hexeberg \textit{et al.} \cite{hexeberg2017ais} propose a novel recursive approach which uses the nearby historical AIS messages around the prediction location to estimate the new position, but it is only for short-term prediction less than 15 minutes.  Furthermore, the method may be sensitive to decision parameters and not suitable for data sparse region (i.e. open sea, at which even nearby messages could be very far). Another work presented in \cite{dalsnes2018neighbor} may have similar constraints. There are also some earlier works, such as neural network approaches\cite{xu2012novel, zissis2016real, zandipour2008probabilistic}, Kalman Filtering approaches \cite{perera2012maritime,perera2010ocean, prevost2007extended}, Minor Component Analysis \cite{BZ2005minor}, fuzzy logics \cite{pallotta2013vessel} and kernel density estimation \cite{ristic2008statistical}. 
 
In this paper, considering the characteristics of historical AIS data and vessels' trajectory segments,  we propose a comprehensive approach to attack the problem of modeling a large volume of raw AIS data for path prediction. Particularly, our work consists of three parts. (1)  Historical AIS data contain a considerable amount of motion outliers, which could severely mislead a prediction algorithm and has to be excluded from training data. We developed an efficient algorithm  which is able to automatically detect motion outliers. (2) Raw AIS data lack conciseness and representativeness. We propose a sample representation method, which constructs concise and uniform features from raw AIS data to enhance the effectiveness of data representation and learning. (3) We propose a motion trend ensemble learning algorithm, which combines a group of predictive models corresponding to different motion trend so that the overall predictive capability is much greater. In addition, we also construct an AIS database containing over 100 GB AIS data collected from hundreds of different types of vessels and, based on the database,  we conduct extensive experiments to show the performance improvements of our approach.

The remainder of the paper is  organized as follows: Section II gives a brief introduction to the Automatic Identification System and AIS data challenges.  Section III  presents  the proposed comprehensive approach to learn predictive models from historical AIS trajectory segments. Section IV reports experimental results for a comparison of the proposed method with existing popular approaches, followed by discussions and conclusions in Section V.

\section{Automatic Identification System and AIS Data Characteristics}
\begin{figure}[ht!]
	\centering
	\includegraphics[width=0.48\textwidth]{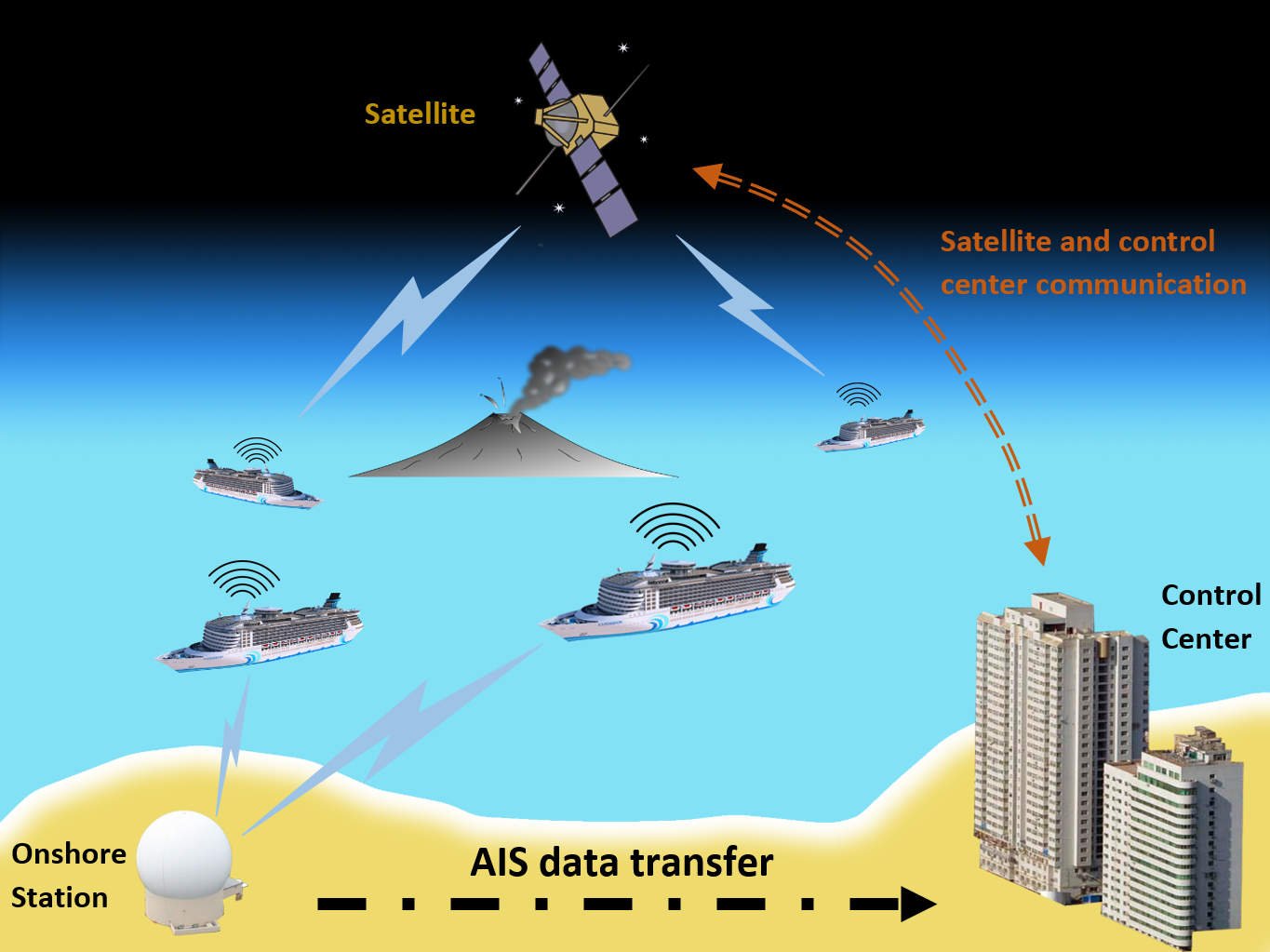}
	\caption{Illustration of Automatic Identification System (AIS).}
	\label{AISSystem}
\end{figure}

The Automatic Identification System (AIS) is a global, autonomous tracking system which  consists of a ship on-board transceiver, ground/satellite station receivers  and vessel traffic services terminals, as illustrated in Figure \ref{AISSystem}.  The onboard Very High Frequencies (VHF) transceiver  broadcasts automatically its host vessel's kinematic information (including ship location, speed, course, heading, rate of turn, destination and estimated arrival time) as well as static information (including ship name, ship MMSI ID, message ID, ship type, ship size, current time) in an AIS message. The  between-message interval is  2 to 10 seconds, depending on the vessel’s speed while underway, and  3 minutes while the  vessel is at anchor.  Meanwhile, each transceiver also receives  AIS messages from other vessels within 20 nautical miles away.  The messages can also be received by  an onshore station, a typical coverage range about 60 nautical miles, or  by a satellite, a coverage range more than 400 nautical miles \cite{clazzer2014impact, recommendation20092169}. These collected AIS data can be transferred in long distance and stored in large volume, and have a big value for maritime data mining and intelligent navigation. 

\begin{figure}[ht!]
	\centering
	\includegraphics[width=0.45\textwidth]{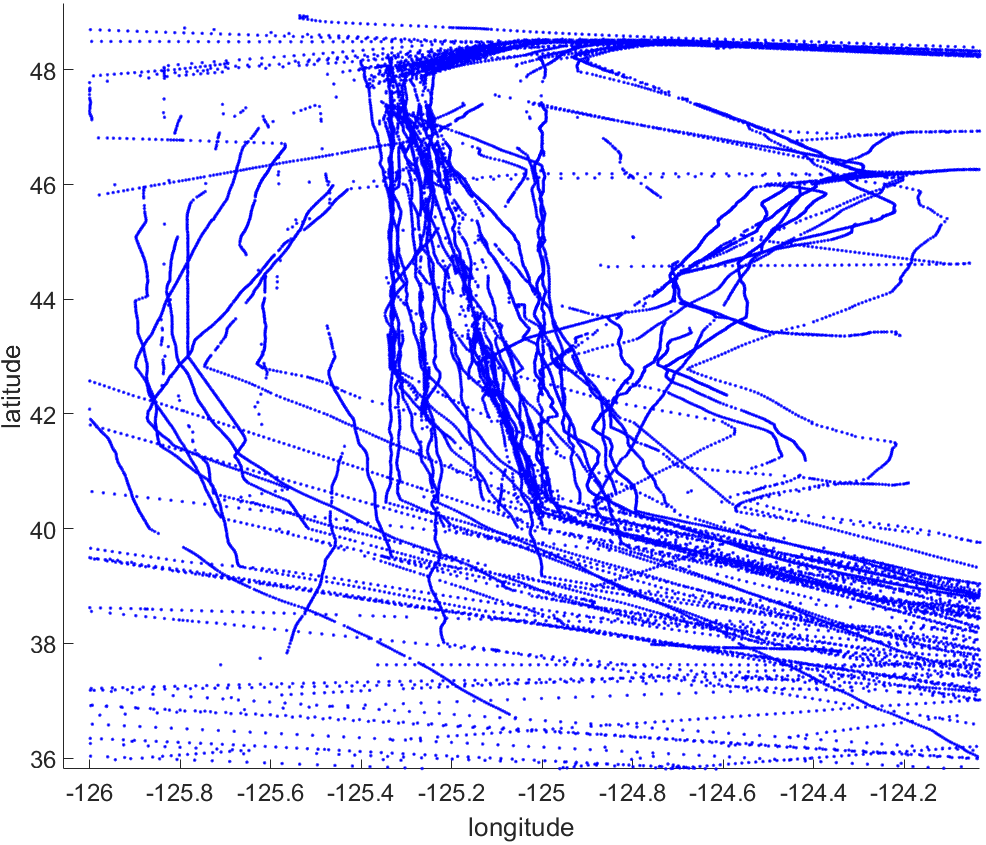}
	\caption{Examples of historical AIS trajectories.}
	\label{AISData}
\end{figure}

However, modeling massive  historical AIS data  for intelligent navigation (e.g path prediction, situation awareness) remains a challenging task. To be more specific, let us see Figure \ref{AISData}, which shows a small part of our AIS database of vessels near the west coast of the USA (more details in Section IV). In the figure, each dot represents an AIS message and each line or curve is an AIS trajectory segments of a vessel. From the figure, we can see that historical AIS trajectories at least have the following issues. (1) The trajectories have a big variance in terms of length, shape, location, and orientation. (2) AIS trajectories frequently have some "abnormal" vessel motion patterns (e.g. wavering, U-turn, self-intersection) which could mislead a learning algorithm and thus reduce the generalization ability. (3) The AIS trajectories  have irregular message frequency, i.e. some trajectories have dense messages sequences but others may have very sparse message sequences, with possible missing data and  erroneous values. Furthermore, each AIS trajectory contains both time-varying features (e.g.  position, speed, whose values change with time) and static features (e.g. vessel size, vessel type, which has a single value for all the time).  These issues, together with the big data volume, pose a big challenge to existing approaches \cite{tu2017exploiting}.  


\section{A Comprehensive Modeling Approach}
As mentioned earlier, our comprehensive modeling approach contains three core parts: the trajectory outliers detection, sample effective feature representation and  motion trend ensemble learning. 

Let us first introduce some notations  which will be used in the following contents.  Suppose an AIS trajectory set contains $N$ trajectories $\{\gamma^{(1)}, \gamma^{(2)}, ..., \gamma^{(N)}\}$\footnote{An AIS trajectory is a series of AIS messages sorted chronologically.}, possibly from different types of vessels and different geographical locations. The $k^{th}$ AIS message in the $n^{th}$ trajectory is denoted by $\gamma_{k}^{(n)} , k=1...{{K}_{n}}$, where $K_n$ is the total number of messages in  the $n^{th}$ trajectory. Note that $\gamma_{k}^{(n)}$ is an AIS message which contains many data entries, such as latitude $\gamma_{k}^{(n)}(\xi)$, longitude $\gamma_{k}^{(n)}(\eta)$, speed over ground (SOG) $\gamma_{k}^{(n)}(sog)$, course over ground (COG) $\gamma_{k}^{(n)}(cog)$, date and time $\gamma_{k}^{(n)}(\tau)$ (For a complete list of all AIS message data entries, please see \cite{harati2007automatic}). We will simply use $\gamma$ for $\gamma^{(n)}_k$ when the meaning is clear in the context.

\subsection{Trajectory Outliers Detection}
AIS trajectories may contain some non-generalizable segments (trajectory outliers) which may be due to some unusual vessel movements (such as avoiding barrier or collision, traffic regulation). If they are not excluded from training data, these outliers could severely mislead a learning algorithm, hence harmful to predictive modeling\footnote{It should be mentioned that it is possible to model the motion outliers specially for anomalous movement prediction, but this is an even more challenging task and out of the scope of this paper. We may leave it as our future work.}.

\begin{figure}[h]
	\centering
	\subfloat[Type I: sharp turning]{
		\includegraphics[width=0.1\textwidth]{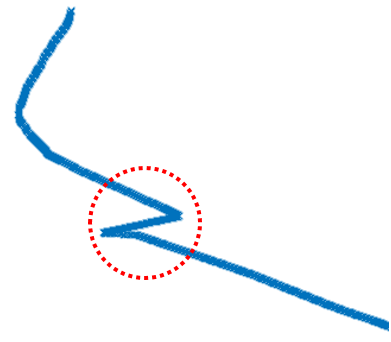}
		\includegraphics[width=0.1\textwidth]{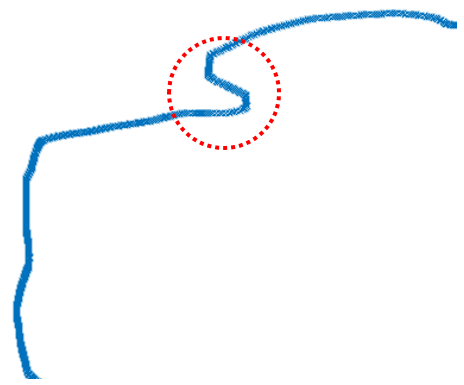}} \hspace{0.2cm}
	\subfloat[Type II: self crossing]{
		\includegraphics[width=0.11\textwidth]{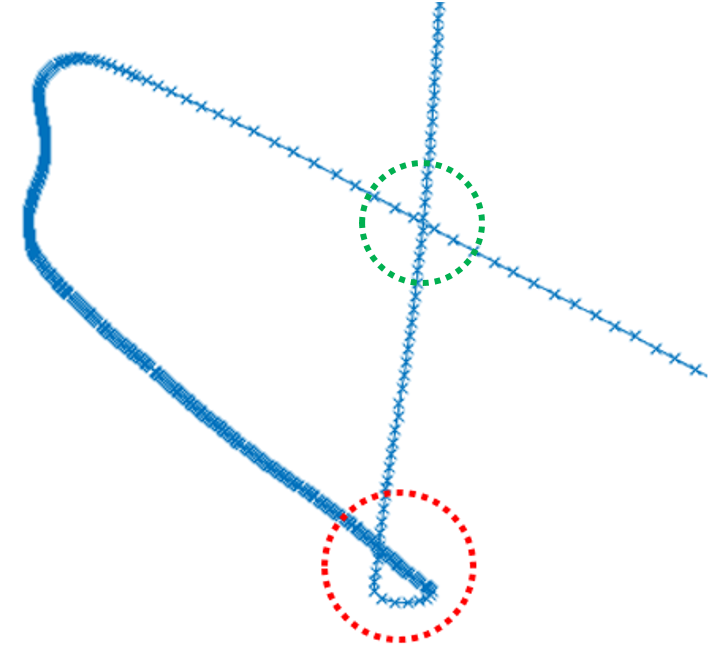}
		\includegraphics[width=0.11\textwidth]{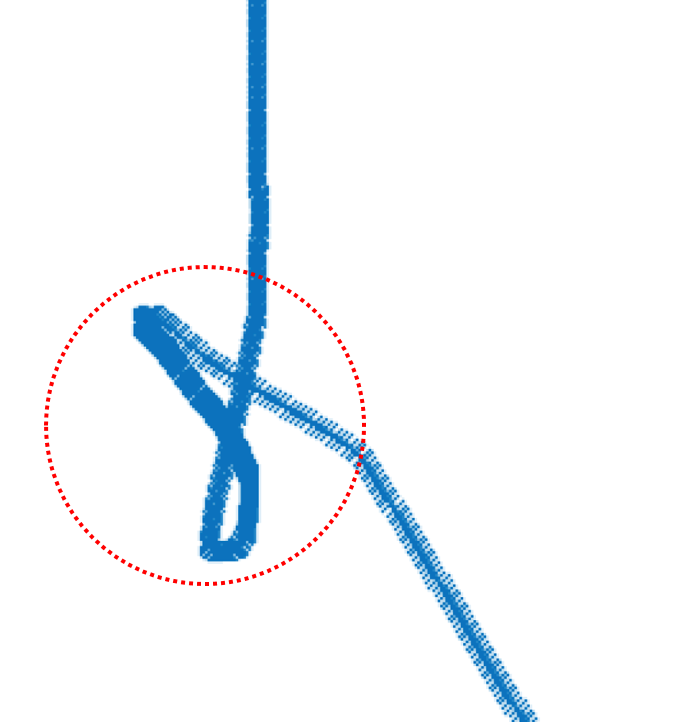}}
	\caption{Examples of anomalous motion patterns, indicated by red dash circles.}
	\label{OutliersMotion}
\end{figure}

The difficulty is that the circumstances of unusual vessel movements may be different from one to another and thus the appearance of the trajectory outliers could also be quite different. Therefore, it is hard to give a concrete detection rule which could be used to identify various unusual trajectory segments. However, fortunately, we found that trajectory outliers are, or can be decomposed into, two basic anomalous motion patterns: sharp turning and self crossing, illustrated in Figure \ref{OutliersMotion} (we call them type I and type II, respectively). If we could accurately locate these anomalous motion patterns,  we could detect the trajectory outliers easily. 

For type I, we make use of the mean-value theorem  to propose an improved Ramer-Douglas-Peucker  (RDP) algorithm \cite{douglas1973algorithms} in Algorithm \ref{TypeI}. The input parameters $d_0$, $\Delta a$ and $\Delta \theta$ are the minimum distance, the slope tolerance between parallel lines and the threshold for sharp turning, respectively. The function $RDP2$ in line 1 finds control points recursively, as implemented in lines from 11 to 28. Then lines 5-7 calculate the angle formed by three neighboring control points (See  \textit{Appendix B} for function $Angle$). If the angle is too small, then the middle control point has a type I anomalous pattern. 

In function $RDP2$, line 11 calculates the slope of the secant of the trajectory curve.  Then function $Distance$ in line 16 computes the distance between point $f_k(\xi,\eta)$ and the line formed by points $f_1(\xi,\eta)$ and $f_K(\xi,\eta)$.  The $Slope$ function in line 17 computes the average  slopes of the front and back lines at the central point, as implemented \textit{Appendix B}. In line 18, if the slope difference between  $a_k$ and $a$ is less than $a_0$ and the distance between $f_k$ and the secant is larger than $d_0$, then distance $d_k$ and the index $k$ are stored. Thereafter, line 23 finds the index of the control point which has minimal or maximal distance and divides the trajectory at these control points. Then each subtrajectory is  inputted to the function $RDP2$ again to find next level control  points recursively. 

\begin{algorithm}
	\caption{Anomalous  Pattern Removal: Type I}
		\label{TypeI}
		\KwIn{AIS trajectory $f$; parameters $\Delta a, d_0, \Delta \theta$}
	\KwOut{Anamlous Pattens Set $\mathcal{B}$}
	$\mathcal{I}=RDP2(f, d_0, \Delta a)$\;
	$K=Length(f), \,\mathcal{I}=\{1, \mathcal{I}, K\}$\;

	$j=0, \,I=Length(\mathcal{I})$\;
	\For{$i=2,...,I-1$}{
		$v_1=f_{\mathcal{I}_{i}}(\xi,\eta)-f_{\mathcal{I}_{i}-1}(\xi,\eta)$\;
		$v_2= f_{\mathcal{I}_{i}}(\xi,\eta)-f_{\mathcal{I}_{i}+1}(\xi,\eta)$\;
		$\theta = Angle(v_1,v_2)$\;
		\If{$\theta < \Delta\theta$}{
			$j=j+1$\;
			$\mathcal{B}_j=\mathcal{I}_i$\;
		}
	}

	\vspace{0.2cm}
	\bf Function $RDP2(f, d_0, \Delta a)$\\
	$a=(f_1(\eta)-f_K(\eta))/(f_1(\xi)-f_K(\xi))$\;
	$j=0, \,K=Length(f)$\;
	$D=\emptyset, \, \mathcal{I}=\emptyset$\;
	\For{$k= 2,3, ..., K-1$}{
		$d_k=Distance(f_k(\xi,\eta), f_1(\xi,\eta), f_K(\xi,\eta))$\;
		$a_k=Slope(f_{k-1}(\xi,\eta), f_k(\xi,\eta), f_{k+1}(\xi,\eta))$\;
		\If{$ |a_k-a|< a_0$ \bf and $d_k>d_0$}{
			$j=j+1$\;
			$D_j=d_k$\;
			$\mathcal{J}_j=k$\;
		}
	}
\If{$D \neq \emptyset$}{
	$i, j=Minmax(D)$\;
	$\mathcal{I}_1=RDP2(f_{1:\mathcal{J}_{i}}, d_0, \Delta a)$\;
	$\mathcal{I}_2=RDP2(f_{\mathcal{J}_{i}:\mathcal{J}_{j}}, d_0, \Delta a)$\;
	$\mathcal{I}_3=RDP2(f_{\mathcal{J}_{j}:K}, d_0, \Delta a)$\;
	$\mathcal{I}=\{i, j, \mathcal{I}_1,\mathcal{I}_2,\mathcal{I}_3\}$\;
	}
	\bf return $\mathcal{I}$
	
\end{algorithm}

\begin{figure}[h]
	\centering
	\subfloat[Target curve]{
		\includegraphics[width=0.15\textwidth]{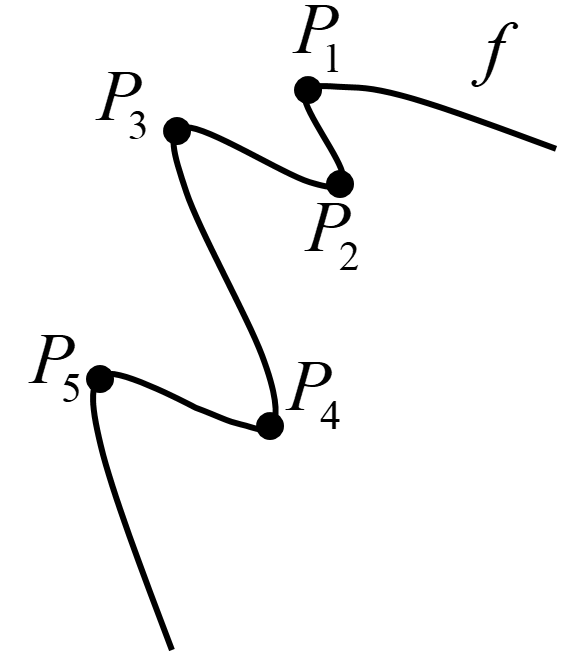}}
	\subfloat[Original RDP]{
		\includegraphics[width=0.15\textwidth]{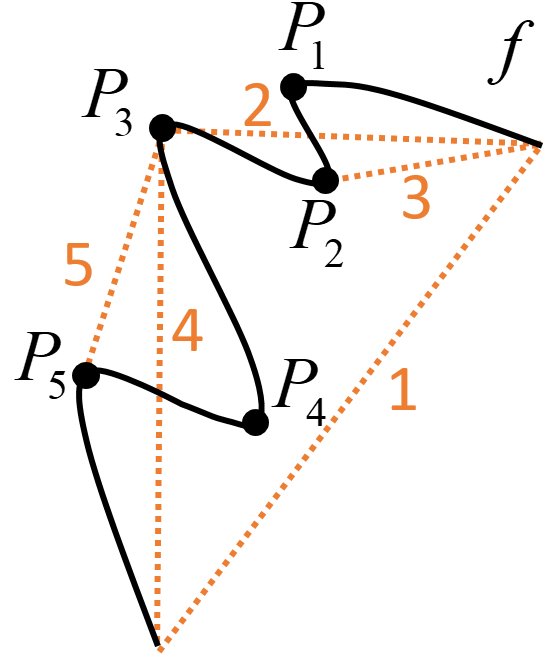}}
	\subfloat[Improved RDP]{
		\includegraphics[width=0.15\textwidth]{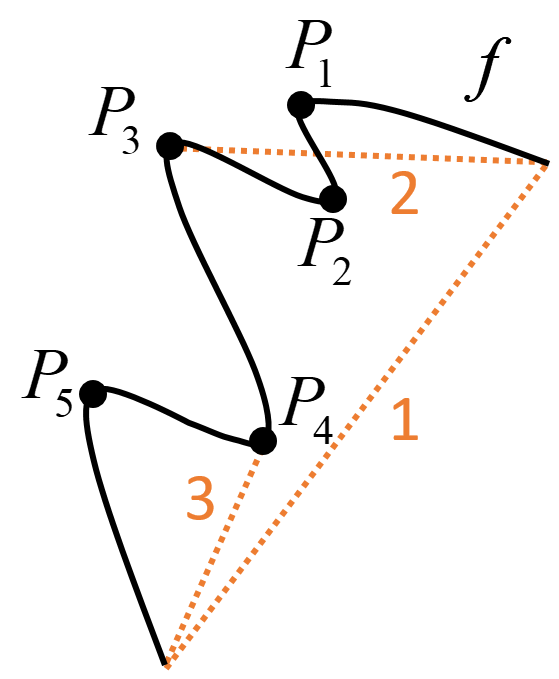}}
	\caption{Comparison of the original RDP algorithm and the improved RDP algorithm. In (b) and (c), the numbers indicate the sequence of recursive function calls and the dash lines are the secants of the corresponding recursive curve segment. }
	\label{RDP}
\end{figure}

The introduction of slope in the improved RDP algorithm enables us to rapidly find  both the nearest and farthest control points  at one time. To show the difference, figure \ref{RDP} gives an example. The original RDP algorithm requires 5 recursive function calls ($P_3\rightarrow P_2\rightarrow P_1\rightarrow P_5\rightarrow P_4$), while the new algorithm requires only 3 recursive function calls ($P_3, P_4\rightarrow P_1, P_2\rightarrow P_5$). Therefore, the new algorithm could reduce recursive function calls significantly, hence speed up the detection process.

For type II anomalous pattern, we need to  find first the intersection location of two line segments in the trajectory and then determine if the intersection is anomalous (There are two intersections in the third figure of Figure \ref{OutliersMotion}, but the one in the green circle could be a normal change of sailing course.). We adopt the following algorithm which is widely used in computer graphics \cite{kirk2012graphics}. More specifically, for two line segments, say $ {{P_1}{P_2}} $ and $ {{P_3}{P_4}} $, we solve the following equations to find $\alpha$ and $\beta$:
\begin{equation}
\left\{ \begin{array}{l}
{P^*} = P_1 + \alpha (P_2 - P_1)\\
{P^*} = P_3 + \beta (P_4 - P_3)
\end{array} \right.
\end{equation}
Two line segments have an intersection if and only if $\alpha \in [0, 1]$ and $\beta \in [0, 1]$. The intersection point $P^*$ can be computed from either equation. Then we examine the trajectory length cut by the intersection point (i.e. start from $P^*$ and end at $P^*$) and mark it as an anomalous pattern if the length is shorter than a predefined parameter, since a small loop usually corresponds to an abnormal movement, as shown in Figure \ref{OutliersMotion} (b).

\subsection{Trajectory Sample Representation}
Most of the existing works learn on original AIS trajectory data directly \cite{tu2017exploiting}, which lack conciseness and effectiveness because the raw data are redundant and contain both static and temporal components with an irregular time stamp. Here, we propose a novel way to transform the original data into uniform feature vectors, an effective and algorithm-friendly representation, by three steps described as follows.

The \textit{\textbf{first step}} is to transform the irregular temporal component of a trajectory into uniform samples of equal length\footnote{Sample could be equal spanning time, equal spanning distance length or equal message number.}. Here we follow the way described in \cite{mao2018automatic} to make overlapped training samples from  the trajectory, because the method could make sure that most of the raw data could be translated into the training data. After this, let $\mathcal{S}$ be the sample set and $\mathcal{T}$ be the corresponding training target set, in a one-by-one correspondence. Here each sample $s_i \in \mathcal{S}$ is a small fragment of the trajectory and its target, $f_t \in \mathcal{T}$, is an AIS message, which is $\tau_t$ time (the prediction time) later after the last message in $s_i$.


Comparing to the vastness of the earth surface, the trajectories are very sparse in most water regions except for near shore port areas. So it could easily result in an overfitting model if a machine learning algorithm is applied simply to the original GPS positions in the AIS data.  To avoid this problem, our \textit{\textbf{second step}} of obtaining effective training samples is to map the GPS positions to a local coordinate system, which is \textbf{\textit{homeomorphic}} to the original GPS coordinate space but the trajectory data distributes much densely, so we could learn a model in a more efficient way in this new coordinate system. It is worth mentioning that to guarantee the homeomorphism, the mapping has to be invertible. Under this condition,   the prediction results given by the model during testing stage could be mapped back to earth GPS positions. Otherwise, the mapping is meaningless and the model would be useless. 

To this end, we first define the sample direction  $\theta$ to be the smallest angle between the tangent line and horizontal direction, i.e. for a sample $s \in \mathcal{S}$ from the sample set
\begin{equation}
\theta =\text{sign}({{v}})\arccos \frac{v^{T}u}{\left\| v \right\|\cdot \left\| u \right\|}
\end{equation}
where vector ${v=(\Delta f(\xi), \Delta f(\eta))}$ is the latitude and longitude coordinate difference between the first two messages in $s$ and $u=(1,0)$ is a horizontal unit vector.  $\text{sign}(v)$ is a sign function, which gives 1 if $v(2)$ is positive and -1 otherwise.
Thereafter we compute a mapping  matrix $\Gamma$ as
\begin{equation}
\label{MappingMatrix}
\Gamma \triangleq \left[ \begin{matrix}
\alpha  & \beta  & \alpha {{\xi}_{0}}+\beta {{\eta}_{0}}  \\
-\beta  & \alpha  & \alpha {{\eta}_{0}}-\beta {{\xi}_{0}}  \\
0 & 0 & 1  \\
\end{matrix} \right]
\end{equation}
where $\alpha=\cos(\theta)$ and $\beta=\sin(\theta)$. $\xi_0$ and $\eta_0$ are the latitude and longitude, respectively, of the first AIS message in $s$.  Note that since $\alpha$ and $\beta$ cannot simultaneously be zero for any value of $\theta$,  the matrix $\Gamma$ is always invertible. Finally, we can apply $\Gamma$ to the latitude and longitude coordinates of each AIS message in $s$ (together with its training target) to map them from the real geographical coordinates into a local coordinate system\footnote{We should point out that at a first glance, this operation seemingly does not make much sense, because the mapping just shifts the sample to the origin point (0, 0) and then rotates it to align to the horizontal direction rightward. But, in fact, the mapping is a very effective and important operation to overcome the model overfitting issue. It gathers those sparsely distributed trajectory samples from vast GPS coordinate space and put them into a dense region in a local space, which is \textbf{\textit{homeomorphic}} to the earth GPS space. Meanwhile the mapping keeps the relative spatial relationship (hence the motion trend to be learned) between the sample and its training target.}. In the prediction stage, we need to perform a reverse operation to map the prediction value back to real geographical coordinates, i.e. latitude and longitude. Since $\Gamma$ is invertible, the only thing needs to do is  multiplying $\Gamma ^{-1}$ to the model prediction output.

After transformed by $\Gamma$,  let the new sample be $\bar s$ and its $l$ AIS messages be $(\bar f_1, \bar f_2, ..., \bar f_l)$. We make use of both kinematic and static information in the AIS messages to design a group of representation features, which are able to reflect the instantaneous motion status (i.e. position, velocity, acceleration) during the short period of the $l$ messages, and meanwhile overcome the problem of irregular message intervals. The features are as follows.
\begin{itemize}
\item Latitude features. Latitude coordinates of  the  first, middle and  last two AIS messages 
	$$ \bar f_1(\xi), \bar f_2(\xi), \bar f_{l/2}(\xi),  \bar f_{l/2+1}(\xi),  \bar f_{l-1}(\xi),  \bar f_l(\xi)$$ 
 The reason is that the beginning, middle and end  points are the three most important instantaneous positions that record the vessels motion dynamics. Besides,  the  average of the latitudes in the first half sample and the second half sample, respectively, also  reflect the vessel's overall sailing trend during that period
	$$ \frac{2}{l}\sum\limits_{k=1}^{l/2}{{{\bar f}_{k}}(\xi)}, \quad \frac{2}{l}\sum\limits_{k=l/2+1}^{l}{{{\bar f}_{k}}(\xi)}$$
	
\item Longitude features. Similarly, the longitude coordinates of the first, middle  and  last two AIS messages 
	$$\bar f_1(\eta), \bar f_2(\eta),  \bar f_{l/2}(\eta), \bar f_{l/2+1}(\eta), \bar f_{l-1}(\eta), \bar f_l(\eta)$$  
	and the averages of the longitudes of the front half sample and  second half sample  $$\frac{2}{l}\sum\limits_{k=1}^{l/2}{{{\bar f}_{k}}(\eta)}, \quad  \frac{2}{l}\sum\limits_{k=l/2+1}^{l}{{{\bar f}_{k}}(\eta)}$$

\item Velocity features. At any moment, the motion direction of a vessel can be decomposed into two components: latitude   and longitude .  Since a vessel's future position is closely related to the latest motion status, we compute two instantaneous velocity values and one mean velocity value along the two directions. So the latitude velocity features are
	$$\frac{\Delta {{\bar f}_{l-1}}(\xi)}{\Delta {{\tau}_{l-1}}},\frac{\Delta {{\bar f}_{l}}(\xi)}{\Delta {{\tau}_{l}}},\frac{1}{l}\sum\limits_{k=1}^{l}{\frac{\Delta {{\bar f}_{k}}(\xi)}{\Delta {{\tau}_{k}}}}$$ where $\Delta \bar f_k(\xi)$ is the difference $\bar  f_k(\xi)-  \bar f_{k-1}(\xi)$ and $\Delta \tau_k$ is the time interval between the $k^{th}$ AIS message and the $(k-1)^{th}$ AIS message. Correspondingly, the longitude velocity features are $$\frac{\Delta {{\bar f}_{l-1}}(\eta)}{\Delta {{\tau}_{l-1}}},\frac{\Delta {{\bar f}_{l}}(\eta)}{\Delta {{\tau}_{l}}},\frac{1}{l}\sum\limits_{k=1}^{l}{\frac{\Delta {{\bar f}_{k}}(\eta)}{\Delta {{\tau}_{k}}}}$$

\item Features of speed over ground (SOG), the acceleration and sailing direction (course over ground, COG) of the vessel. $$\frac{1}{k}\sum\limits_{i=l-k}^{l}{{{\bar f}_{i}}\left( sog \right)}, \frac{1}{k}\sum\limits_{i=l-k}^{l}{\frac{\Delta {{\bar f}_{i}}\left( sog \right)}{\Delta {{\tau}_{i}}}}, \frac{1}{k}\sum\limits_{i=l-k}^{l}{{{\bar f}_{i}}(cog)}$$
	where $\Delta  \bar f_k(sog)$ is $ \bar f_k(sog)- \bar f_{k-1}(sog)$ .   The first term is the average of SOG  and the second term is the average of SOG change rate (acceleration). The third term is the average of COG. Here only the last $k$  AIS messages are utilized to define the features. 
	\item Other static information in AIS messages, including vessel size, type and drought, since they also deliver important information of the vessel motion characteristics.
\end{itemize}

So far, for the sample $\bar s$,  we could construct its a feature vector, denoted as $x$, to summarize  the instantaneous motion dynamics of the vessel by concatenating these feature values together. Comparing with raw AIS data entries, as have been used in some existing work such as \cite{ristic2008statistical, xu2012novel, zissis2015real}, the new features  are  more concise, representative and effective. Importantly, even if the same length  and AIS messages frequency may vary considerably, the feature vector maintains a fixed length and representativeness.

 \begin{figure*}[ht!]
	\centering
	\includegraphics[width=0.7\textwidth]{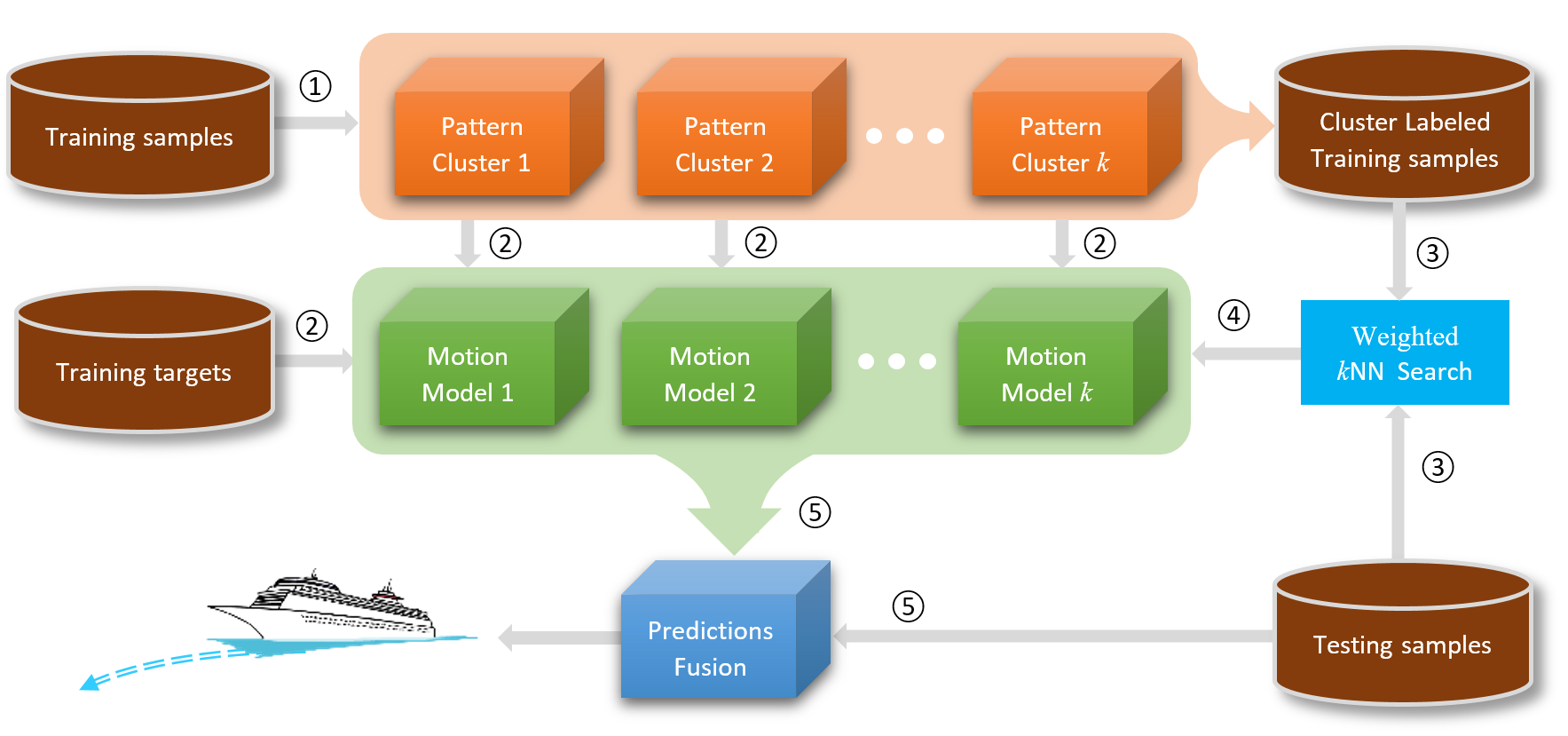}
	\caption{An ensemble learning algorithm. The numbers in the circles indicate learning procedures.}
	\label{EnsFrame}
\end{figure*}



\subsection{Motion Trend Ensemble Learning}
We have computed a feature vector to represent each sample. Putting all samples' feature vectors together we make a training data set denoted by $(\mathcal{X}, \mathcal{Y})$,  where $\mathcal{X}=\{x_1, x_2, ..., x_N\}$ is the training samples set and $\mathcal{Y}=\{y_1, y_2, ..., y_N\}$ is the corresponding training target set. Each $y_i$ is a pair of transformed latitude and longitude $(\bar f_t(\xi), \bar f_t(\eta))_i$. Figure \ref{EnsFrame} displays the architecture of the proposed ensemble learning algorithm. As indicated by  the circled numbers in the figure, the training samples are  first clustered into $k$ clusters, say $\mathcal{X}_1, \mathcal{X}_2, ..., \mathcal{X}_k$, by an unsupervised learning algorithm  (here we use $k$means for simplicity, but other unsupervised learning algorithms are equally applicable). The main purpose of this step is to reduce the diversity and imbalance (hence the modeling difficulty) of vessel movements for a modeling algorithm. The resultant clusters have two usages: the first one is to train a predictive model for each cluster learning stage (circled 2) and the second one is to perform model selection and fusion in  testing stage (circled 4 \& 5). For the first usage, given a pattern cluster $\mathcal{X}_i \subset \mathcal{X}$, the  corresponding training target set $\mathcal{Y}_i$ could be collected from $\mathcal{Y}$.  Then for each pair of $(\mathcal{X}_i, \mathcal{Y}_i)$, we could build a  motion model by training a supervised learning algorithm,  such as neural networks, least square support vector machine, etc. In our current implementation, we use the extreme learning machine (ELM) \cite{huang2006universal, huang2012extreme} which is a special type of multilayer neural networks and is very efficient. 
 
 At the testing stage, the clusters 
 $\mathcal{X}_1, \mathcal{X}_2, ..., \mathcal{X}_k$ are assigned to labels 1, 2, ..., $k$, respectively. All samples in the same cluster share a cluster label. Given a testing sample $x$, the (weighted) $k$-nearest-neighbor ($k$NN) algorithm could be adopted to find its closest $K$ neighboring from the labeled clusters (circled 3). Each nearest-neighbor sample is fed to the model corresponding to the labeled cluster to get its prediction, hence the prediction error regarding to original training target in $\mathcal{Y}$. The first $r (r\le k)$ models with lowest prediction error and the associated nearest-neighbors (a subset of the $K$ nearest-neighbor samples) are selected (circled 4). The final prediction result of $x$ can be obtained by the following fusion scheme 
 \begin{equation}
 	y=({\sum\limits_{i=1}^{r}{{{w}_{i}}{{{\hat{y}}}_{i}}}})/{\sum\limits_{i=1}^{r}{{{w}_{i}}}}\;
 \end{equation}
 where ${\hat y}_i$ is the prediction result of $x$ given by the $i^{th}$ selected model and ${{w}_{i}}=\exp \left( {-e_{i}^{2}}/{2{{\sigma }^{2}}}\; \right)$. $e_i$ is the average prediction  error of selected model $i$ on the nearest-neighbors.

 The advantages of the proposed ensemble learning are as follows. First, modeling each pattern cluster separately can achieve better prediction accuracy, because the diversity and imbalance,  hence the modeling difficulty,  within each cluster are significantly reduced. Second, training an  algorithm with a smaller cluster usually saves much computation cost, because the computational complexity of most machine learning algorithms is polynomial with respect to the number of training samples.  Finally, it is more flexible to model each motion pattern cluster individually, because each model can be adjusted without any influence on other models' performance. This is especially important when some new samples are added or part of samples are removed. In this case, only the related model needs to be updated. Furthermore, the fusion of multiple models with similar movements tends to be stabler than a single one.

\section{Experimental Results}
In this section, we conduct experiments to compare the proposed approach with several popular path prediction methods on a real-world dataset, which constitutes AIS messages from both ground stations and satellite stations.  
\subsection{AIS Database and Experimental Dataset}
We implemented a web data crawler which continuously crawling real AIS data from the Internet to build an  AIS database  for vessel path prediction and maritime data mining.  The database is  about 100GB and  is from the international data provider Marine Cadastre \footnote{http://marinecadastre.gov}, containing AIS messages of both ground stations and satellite stations. 

\begin{figure}[H]
	\centering
	\includegraphics[width=0.24\textwidth]{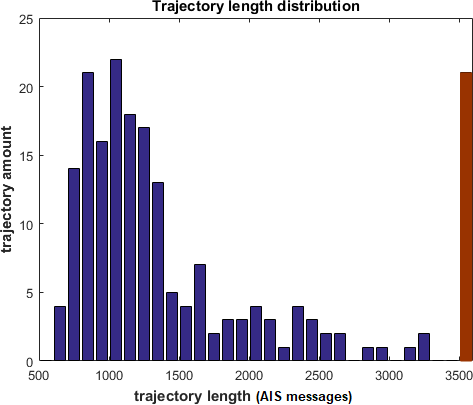}
	\includegraphics[width=0.24\textwidth]{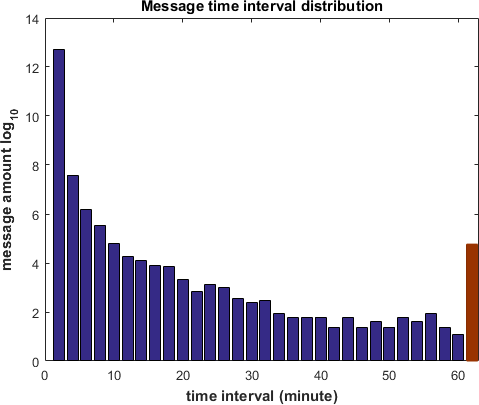}
	\caption{Trajectory length distribution (left) and message time interval distribution (right). }
	\label{TrajectoryStat}
\end{figure}

For our experimental purpose, we only use part of the database to construct an experimental dataset, because the whole database is too huge for a learning algorithm running on a single computer. From the database, 200   trajectories  from 180  different vessels (including cargo, tank, towing \& tug vessel, fishing boat, lifeboat, etc) are extracted to form a dataset\footnote{Due to copyright concern, the dataset is only available  upon request.} and then the preprocessing algorithm in \textit{Appendix A} is applied to remove some apparent issues.  Details of the trajectory extraction process and related information of the database can be found in \cite{mao2018automatic}. 
Fig. \ref{TrajectoryStat} shows the trajectory length distribution and message time interval distribution in the dataset. In each figure, we  display the overall amount for all abscissa quantity that is out of the range (larger than the right-most abscissa value) in the last bin (brown color).   From these figures, we can see the irregular trajectory lengths and between-message time intervals. Note that the intervals distribute in a rather broad range, i.e. from minutes to hours long. Most predictive approaches (for static vector or time series ) requires either fixed sample length or fixed step size \cite{bishop2006pattern, weigend2018time}, so it will be very low efficiency even infeasible to training algorithms directly using real-world raw AIS data. 

\subsection{Single Path Prediction Simulation}
\begin{figure}[h]
	\centering
	\subfloat{
		\includegraphics[height=5cm]{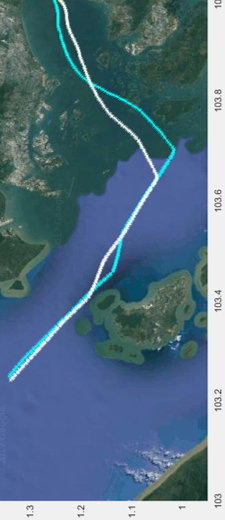}}
	\subfloat{
		\includegraphics[height=5cm]{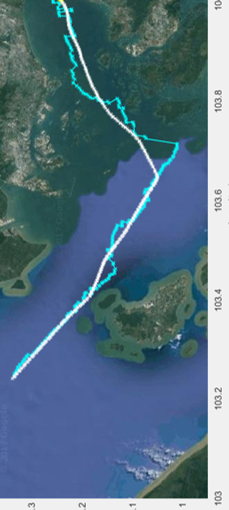}}
	\subfloat{
		\includegraphics[height=5cm]{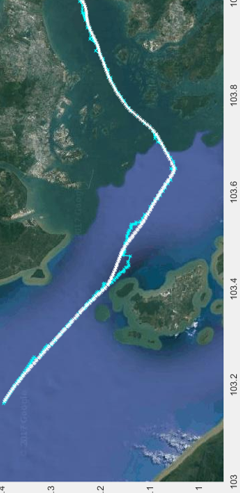}}
	\caption{Single path simulation results. From left to right: linear , SOG-COG  and the proposed method. The white line in each map is the true sailing trajectory and the blue line is the predicted result.}
	\label{SinglePrediction}
\end{figure}

To demonstrate why machine learning based path prediction outperforms straightforward linear projection and velocity calculation, we first  run an experiment for a single path prediction to compare  the capability of linear projection method, SOG-COG estimate method and our  method. Linear projection method performs a forward linear interpolation to predict next time position using the latest three AIS messages.  SOG-COG estimate method computes the next time position by multiplying SOG with time gap and project it to current COG direction. While these two methods are  most basic and straightforward, they are very illustrative and intuitive to show the difference and importance of machine learning based approaches. Figure \ref{SinglePrediction} displays the simulation results for 30 minutes prediction (The smoother the predicted path is and the closer the  predicted path to the true sailing trajectory, the better the result is).

From these results, we can see that the linear prediction method is able to output a smooth predicted path, but the deviation is very big.  SOG-COG estimate method could follow true sailing trajectory more closely, but its results perturb significantly because of the fluctuations in instantaneous speed and course values.  The proposed method is able to give a much more accurate prediction. The reason  is that linear and SOG-COG methods are based on a vessel's instantaneous motion state (current or one time step back) and thus the predicted future path is also "short-sight" and inaccurate.  In the contrast, the proposed method learns a group of models using a large amount of historical AIS trajectories, which contain various complex motion patterns. As a result,  the learned models \textit{remember} all types of motion patterns and could \textit{predict}  future vessel motion trends earlier.
 
%
%

\subsection{Quantitative Comparison Results}
In this section, we conduct  experiments on the AIS dataset to compare quantitatively the proposed approach  with Gaussian Process Regression (GPR) , Least Square Support Vector Machine (LS-SVM), Multilayer Perceptron (MLP)  and Gaussian Mixture Models (GMM). We choose these algorithms as the baseline because they are popular for path prediction and able to achieve state of the art results  \cite{zhou2010lssvm, kowalska2012maritime, xu2012novel, zissis2015real, wiest2012probabilistic}\footnote{We also run Kalman filtering based prediction algorithms, but their prediction errors are very large for our tasks. The reason  may be that Kalman filtering is suitable for short-term prediction, usually the next time stamp within seconds length, but here we make much longer  predictions on a low-quality dataset, from 15 minutes to 1 hour.}. The Original Extreme Learning Machine (ELM) algorithm  (applying directly to raw) is also included as a baseline algorithm to show the effectiveness of our comprehensive approach. 

We perform multi-step predictions, i.e. four different future time steps (15, 30, 45 and 60 minutes ahead) to compare the capability and flexibility of these path methods. Experiments are run  in a 10-fold cross validation way on the dataset and the final prediction error is the average of the 10-fold results.  The prediction error is measured by the geographical distance (computed by the $GeoDistance$ function in Appendix A) between the ground truth and the predicted result.   The parameters of the baseline algorithms are tuned by cross validation to produce the lowest errors.  The experiments are implemented with MATLAB 2015a and conducted on a computer with 2.0 GHz dual-core Intel CPU and 16 GB RAM memory. 
\subsubsection{Full dataset experiments}
We first conduct experiments on the whole dataset to demonstrate the capability of each algorithm to learn complex motion patterns from a mixture of  a variety of trajectories. The results are shown in Table \ref{FullDataSetExpMean}.  The first column  is the prediction time (unit: minute).
\begin{table}[h]
	\centering  
	\footnotesize
	\caption{Prediction error on the whole dataset (time unit: minute; error unit: nautical mile).}
	\label{FullDataSetExpMean}
	\begin{tabular}{ c c c c c c c}
		\toprule\hline
		               Time    & ELM   & LS-SVM   & MLP     & GMM  & GPR     & Ours  \\
		\cmidrule(r){2-7}
							15 & 1.35  &  1.33  &  3.13 &   1.77 & 2.19 &\textbf{ 0.53} \\
		\cmidrule(r){2-7}
							30 & 3.49 &   3.55  &  5.42  &  5.09 &   5.19 &\textbf{ 1.31}  \\
		\cmidrule(r){2-7}
							45 & 5.85  &  5.90  &  8.03  &  9.44 &   9.89 & \textbf{ 2.34 }\\
		\cmidrule(r){2-7}
							60 & 8.48  &  8.60 &  10.92 &  14.27 &  15.07 &\textbf{ 3.57}\\ \hline\bottomrule
	\end{tabular}
\end{table}

For longer  time prediction, the motion patterns will be more complex and not necessarily Gaussian distribution. So we can see that the performance of  GMM decays dramatically for long time path prediction. GPR and MLP also have big prediction errors. LS-SVM and ELM produce relatively better results for both short-term 15 minutes prediction and long-term 60 minutes prediction, because of their strong regression ability to learn an arbitrary complex continuous function from a set of samples.  In all cases, the proposed method is able to achieve much lower prediction errors. Benefiting from sample representation and ensemble learning, the proposed approach is able to utilize a group of different models to exploit the big amount of historical data to learn  complex motion patterns in a concise feature space. So even the prediction is a much challenging task that includes different types of vessels' trajectories, it still can learn and predict different vessels' motion pattern more accurately. 

Table \ref{FullDataSetExpSTD} shows the standard deviation of the prediction results for each algorithm. The results show that the variance of the proposed approach is  smaller  than or comparable with other algorithms. 
\begin{table}[h]
	\centering  
	\footnotesize
	\caption{Standard deviation of prediction error on the whole dataset (time unit: minute; error unit: nautical mile).}
	\label{FullDataSetExpSTD}
	\begin{tabular}{ c c c c c c c}
		\toprule\hline

		      Time    & ELM   & LS-SVM   & MLP     & GMM  & GPR     & Ours           \\
		\cmidrule(r){2-7}
		15 & 0.63 & 1.26 & 3.66 & 3.46  & 2.56 & \textbf{0.25} \\
		\cmidrule(r){2-7}
		30 & 1.25 & 2.61 & 4.15 & 5.41 & 3.63 & \textbf{1.06} \\
		\cmidrule(r){2-7}
		45 & 2.30 & 3.86 & 4.99 & 7.44 & 5.20 & \textbf{1.85} \\
		\cmidrule(r){2-7}
		60 & \textbf{2.80} & 5.02 & 6.25 & 8.23 & 7.27 & 2.87 \\ \hline\bottomrule
	\end{tabular}
\end{table}

\subsubsection{Side information experiments}
One advantage of the proposed comprehensive approach is that it could make use of some side information, i.e. the contextual information that is not directly included in the AIS data. This has rarely been utilized by previous methods \cite{tu2017exploiting}.  Such information can be collected without costing much effort and is helpful to improve path prediction results. For example,  the same type of vessels with a similar voyage are more likely to have similar motion characteristics and trajectory patterns. Here we investigate the influence of vessel type information and geographical location information.

\begin{table}[h]
	\centering  
	\footnotesize
	\caption{Vessel types and trajectories }
	\label{ShipType}
	\begin{tabular}{ c c c c }
		\toprule\hline
		 Towing vessel & Tug vessel   & Cargo  vessel   & Tanker vessel\\
		18 &42 & 92 & 38 \\ \hline\bottomrule
	\end{tabular}
\end{table}

To include static information, the trajectories in the dataset are divided into groups according to their types and four types of vessels are used: towing vessels, tug vessels, cargo vessels, and tanker vessels. Their corresponding  number of trajectories  are listed in Table \ref{ShipType}. For contextual information, we divide the trajectories into different groups based on their geographical coordinates and finally choose 4 groups, each one containing trajectories in the same geographical region and generally following a  similar traffic route. Then the group number is added directly to each sample feature vectors and the prediction algorithms are run in a similar setting as previous experiments. The experimental results are shown in Table \ref{StaticInfoError} and  Table \ref{GeoInfoError}.  
\begin{table}[h]
	\centering  
	\footnotesize
	\caption{Mean prediction error of including static information (time unit: minute; error unit: nautical mile).}
	\label{StaticInfoError}
	\begin{tabular}{ c c c c c c c}
		\toprule\hline
		Time        & ELM   & LS-SVM   & MLP     & GMM  & GPR     & Ours \\
		
		\cmidrule(r){2-7}
		15 & 1.33  &  1.31  &  3.11 &   1.75  &  2.18 &\textbf{ 0.45} \\
		\cmidrule(r){2-7}
		30 & 3.44 &   3.53  &  5.39 &   5.06  &  5.17 &\textbf{ 1.03}  \\
		\cmidrule(r){2-7}
		45 &5.84  &  5.86  &  7.98 &   9.45 &   9.88 & \textbf{ 2.05 }\\
		\cmidrule(r){2-7}
		60 &8.45  &  8.56 &  10.88 &  14.28 &  15.08 &\textbf{ 3.15}\\
		
	 \hline\bottomrule
	\end{tabular}
\end{table}

\begin{table}[h]
	\centering  
	\footnotesize
	\caption{Mean prediction error of including contextual information (time unit: minute; error unit: nautical mile).}
	\label{GeoInfoError}
	\begin{tabular}{ c c c c c c c}
		\toprule\hline
		Time         & ELM   & LS-SVM   & MLP     & GMM  & GPR     & Ours            \\
			\cmidrule(r){2-7}
	15 & 1.30  &  1.32  &  3.10  &  1.74 &   2.17 &\textbf{ 0.37} \\
	\cmidrule(r){2-7}
	30 & 3.45  &  3.53  &  5.37  &  5.07 &   5.14 &\textbf{ 0.89}  \\
	\cmidrule(r){2-7}
	45 & 5.80  &  5.89  &  7.98  &  9.46  &  9.88& \textbf{ 1.91 }\\
	\cmidrule(r){2-7}
	60 & 8.47  &  8.58  & 10.89 &  14.29 &  15.10 &\textbf{ 3.10}\\	 \hline\bottomrule

	\end{tabular}
\end{table}

Comparing the results with Table \ref{FullDataSetExpMean}, we can see that while most of the baseline algorithms have some very slight improvements, the prediction error reduction of the proposed algorithm is much larger.  While side information is available, the collected massive historical trajectories could be clustered into more homogeneous groups according to the side information and within each group, the trajectory variance will be significantly smaller. In this case, the ensemble learning difficulty and complexity is reduced and the prediction performance is enhanced. A potential application of including the side information into path prediction is that different traffic management strategies could be applied to a different type of vessels or  different traffic contexts in large modern port regions (e.g. Singapore port \cite{weng2012vessel}).

\subsubsection{Time cost comparison}
We also conduct experiments to compare the time cost of all the algorithms. Each algorithm runs 5 times on a subset of the dataset (20 trajectories) and their average time costs are shown in Table \ref{TimeCost}. 
\begin{table}[h]
	\centering  
	\footnotesize
	\caption{Time cost comparison (Unit: Minute)}
	\label{TimeCost}
	\begin{tabular}{ c c c c c c}
		\toprule\hline
		  ELM   & LS-SVM   & MLP     & GMM  & GPR     & AAEL   \\
		\cmidrule(r){1-6}
		3.01 & 3.61 & 110.61 & 29.01 & 40.95 & 1.15\\
 \hline\bottomrule
	\end{tabular}
\end{table}
From this table, we can see that  the proposed approach has a much smaller time  cost than others because it is only trained once on the training data and then makes predictions on testing data requiring no further model update.  MLP and GPR cost much longer time than others because GPR needs to compute a large matrix inverse during training and MLP needs to be trained repeatedly (epoch by epoch) to get converged.  It should be mentioned that the time costs here include both the training and testing phase on a group of trajectories. In real applications, an algorithm usually only needs to make a prediction for \textit{one} sample at each time stamp, so the time cost of the proposed algorithm will be much less and thus meet the real-time requirements in real applications.
\section{Discussions and Conclusions}




Traditional path prediction methods generally based on  single  trajectory learning, such as Kalman filtering \cite{perera2012maritime, perera2010ocean, ra2006real, prevost2007extended}, neural network \cite{khan2005ship, xu2012novel, zissis2015real}, in which a learning model (e.g. Kalman Filtering) is built for a target   vessel and is updated periodically when receiving new data. While for the purpose of vessel collision avoidance, there are  some  drawbacks of this type of approaches. (1) The trained model is associated with the target vessel and thus cannot be used to predict other vessels' future position. This will be highly inconvenient if there are many vessels to be predicted at the same time, such as Singapore Port which is a world famous busy port and needs to regulate hundreds of vessels every day.   (2) These methods usually need to be updated continuously in order to use the latest data to adjust the model to reduce prediction error. This dynamic update brings extra computational burden during deployment and  may prolong the response time during navigation.  (3) These on-line updating models are individual trajectory associated and thus cannot make  use of a large volume of historical data  to learn complex motion patterns  (such as learning a common commuting route from a group of historical trajectories). (4) The online learning methods require a considerable amount of past trajectory data from the target vessel to update a model. This is practically inconvenient for  busy ports, e.g. Singapore Port, because every day there are hundreds of new vessels arrival and their past sailing data are quite limited even absent. 

Recently, as the availability of AIS data increases, constructing a path prediction model based on a large amount of historical AIS data becomes a popular topic in the marine intelligence area. However, previous approaches usually assume that the historical AIS data have been sorted into a good form and the path prediction task is restricted to data-rich (port) area. As described in Section II, in real-world applications both the data and the task are much more challenging than the assumptions.

In this paper, we proposed a comprehensive framework for  massive real-world historical AIS data learning to predict various vessels path with different geographical context. Comparing with existing approaches, the proposed approach is  one-shot learning (train only once and no successive update required) and is able to make multiple predictions for a different type of vessels at the same time. The experimental results have demonstrated its effectiveness for handling the diversity, divergence, and imbalance in trajectory data.  However, the proposed approach also has some limitations. One limitation is that it  requires a big amount of historical data to train the models and thus may limit its application to the scenarios where only very limited data are available. The big data volume may also  require more storage space and computational power.  Another limitation is that it has many user parameters which need to be manually tuned in order to obtain good prediction results. Our future work will  focus on further improving the algorithm performance while maximally reducing the number of user-tunable parameters.

\appendices
\section{A Trajectory Preprocessing Algorithm}
The algorithm attempt to resolve four types of data problems in raw AIS data:  SOG error, COG jump,  message absence and message duplication, since these problems are very common in raw AIS data and have significant negative impact upon the performance of general machine learning algorithms. In Algorithm \ref{PreprocAlg}, $s_0$ and $d_0$ are two threshold parameters for abnormal SOG  value detection and $s_{max}$ is the possible speed limit for different types of vessels \cite{faber2012regulated}. The \textit{Length} function performs a trivial operation count the number of messages in the input trajectory. The \textit{Initialization} function truncates  invalid AIS messages  at the front of the trajectory by  repeatedly examining and removing messages with invalid data entries (i.e. position coordinates are out of range or speed is not reasonable), as implemented in line 24-30.  Lin 5 computes time interval between to messages. 
The function $GeoDistance$ in line 10 computes the actual sailing distance between message $k$ and $k-1$. We use  haversine formula \cite{robusto1957cosine} to compute the earth surface distance, as implemented 31-36. 

\begin{algorithm}
	\caption{AIS Data Filtering}
	\label{PreprocAlg}
	\KwIn{AIS trajectory $f$; parameters $s_0, d_0, s_{max}$}
	\KwOut{Filtered trajectory $g$}
	$K=Length(f)$\;
	$f, K=Initialization(f, K, s_{max})$\;
	$g_1=f_1, \,M=1$\;
	\For{$k= 2,3, ..., K$}{
		$\Delta t = f_{k}(t)-f_{k-1}(t)$\;
		\If{$\Delta t =0 $  }{
			\bf continue}
		$g_{M}=f_{k}, \,M=M+1$\;
		$\Delta d = GeoDistance(f_k(x, y), f_{k-1}(x, y))$\;
		\uIf{$f_{k}(sog)<0$ \bf or $ f_{k}(sog)>s_{max}$}{
			$g_{M}(sog)=f_{k-1}(sog)$\;}
		\uElseIf{$| f_{k}(sog)-f_{k-1}(sog)| > s_0$}{
			$d_p=\Delta t \times f_{k-1}(sog)$\;
			\If{$|\Delta d  - d_p|< d_0$}{
				$g_{M}(sog)=f_{k-1}(sog)$\;}
		}
		
		\If{$\Delta t > t_0$ \bf and $\Delta d >f_{k-1}(sog)$}{
			$m=\Delta d /f_{k-1}(sog)$\;
			\If{$m \geq 3$}{
				$g_{M:M+m-1}=Interpolation(f_{k-1}, f_{k}, m)$\;
				$M=M+m-1$\;}
	}}
	$g_{1:M}(cog)=\text{sin}(g_{1:M}(cog))$\;

	\vspace{0.3cm}
	\bf Function $Initialization(f, K, s_{max})$\\
	
	\For{$k<K$}{
		\If{$f_k(x) \notin[-90, 90]$ \bf or $f_k(y) \notin[-180, 180] $ \bf or $f_k(sog) \notin(0, s_{max}] $}{
			$delete \;message \;f_k$\;
			\bf continue\;}
	}
	$K=Length(f)$\;
	\bf return $f, K$\\
	\vspace{0.3cm}
	\bf Function $GeoDistance((x_1,y_1), (x_2, y_2))$\\
	$R=3440  \hfill  //  earth \;radius$\;
	$h_x={\sin^2}(\frac{x_2 - x_1}{2})$\;
	$h_y={\sin^2}(\frac{y_2 - y_1}{2})$\;
	$d=2R\arcsin \left( \sqrt{h_x+\cos ({{x}_{1}})\cos ({{x}_{2}})h_y} \right)$\;
	\bf return $d$
	
	\vspace{0.3cm}
	\bf Function $Interpolation(f_{k}, f_{k-1}, m)$\\
	\For{$i=1, 2, 3,...,m$}{
		${{g}_{i}}={{f}_{k-1}}+\frac{{{f}_{k}}-{{f}_{k-1}}}{m}\times i$}
	
	\bf return $g$
\end{algorithm}
Line 12 uses previous valid one if current SOG is invalid.  Line 13-16 mean that if the SOG changes too abruptly but the  actual sailing distance supports previous speed, then current SOG value will be replace by previous valid one. Line 17-21 mean that if the time interval between the messages is too long but the vessel is not still, then AIS messages are missing and a linear interpolation operation will be adopted for all time series features to fill the path gap, as implemented in the function $Interpolation$ in line 37-40, assuming that the vessel follows a constant speed and course during the gap period. Line 22 converts course over ground (COG) to sine function values to eliminate the big jumps  of degree value around 360.

\section{Functions used in Algorithm 1}
\begin{algorithm}
	\caption{Anomalous  Pattern Removal: Type I}
	\label{TypeI2}
	
	\bf Function $Slope((x_1, y_1), (x_2, y_2), (x_3, y_3))$\\
	$a_1=(y_2-y_1)/(x_2-x_1)$\;
	$a_2=(y_3-y_2)/(x_3-x_2)$\;
	$a=(a_1+a_2)/2$\;
	return $a$
	
	\vspace{0.3cm}
	\bf Function $Angle(v_1, v_2)$\\
	$a=arccos((v_1^Tv_2)/(\Vert v_1\Vert \cdot \Vert v_2 \Vert))*180/\pi$\;
	return $a$
\end{algorithm}


\ifCLASSOPTIONcaptionsoff
  \newpage
\fi

\bibliographystyle{IEEEtran}
\bibliography{refs}

\vfill\eject

\end{document}